\documentclass[letterpaper]{article}
\usepackage{aaai}
\usepackage{times}
\usepackage{helvet}
\usepackage{courier}
\usepackage{amsmath}

\usepackage{quoting}
\usepackage{graphicx}
\usepackage{hyperref}

\begin{document}

\title{Automatic Game Design via Mechanic Generation}
\author{Alexander Zook and Mark O. Riedl\\
School of Interactive Computing, College of Computing\\
Georgia Institute of Technology\\
Atlanta, Georgia, USA\\
\{a.zook, riedl\}@gatech.edu
}

\maketitle
\begin{abstract}
\begin{quote}
Game designs often center on the game mechanics---rules governing the logical evolution of the game.
We seek to develop an intelligent system that generates computer games. 
As first steps towards this goal we present a composable and cross-domain representation for game mechanics that draws from AI planning action representations.
We use a constraint solver to generate mechanics subject to design requirements on the form of those mechanics---what they do in the game.
A planner takes a set of generated mechanics and tests whether those mechanics meet playability requirements---controlling how mechanics function in a game to affect player behavior.
We demonstrate our system by modeling and generating mechanics in a role-playing game, platformer game, and combined role-playing-platformer game.
\end{quote}
\end{abstract}

\section{Introduction}

%% Mark's suggested version
Can an intelligent system design a computer game?
Games are made of mechanics and content. 
Game mechanics are rules governing the logical state changes in the game; they vary widely across game genres \cite{fullerton2008:playcentric,salen2003:rulesplay}.
Content encompasses all non-procedural assets in a game including art, sounds, animations, levels, and maps.
Previous approaches to automated game generation have focused on hand-designed game-specific mechanics, generating content for fixed sets of hard-coded mechanics, or AI design critics that do not generate games.
As a result, most prior work on game generation revolves around selecting and assembling components from human provided knowledge and content.

In this paper we explore synthesis of game mechanics from low-level game engine primitives related to checking and updating game variables.
Choosing the right mechanics for a game involves deciding what players should be allowed to do (and when) and deciding how the game mechanics should function to achieve these goals for player actions.
By automatically designing game mechanics, an intelligent system can create games unique to each player, generate creative solutions to design problems humans may have not been able to conceive, create games across many domains, or recombine existing mechanics into new game genres.
%[[benefits: domain-independent, find solutions that human designers might not have considered, recombination, etc., do things other people haven’t been able to do.]]

Mechanic generation is \textit{de novo} synthesis of game agent actions given knowledge of a game domain.
We address mechanic generation with a representation for composable game mechanics and accompanying techniques to generate mechanics subject to playability and design requirements.
%
% contributions
Generated mechanics take the form of planning operators in a representation specialized to game mechanics.
Our mechanic design technique is a ``generate-and-test'' process: (1)~generating mechanics that meet design requirements on form and (2)~testing mechanics to ensure they meet playability requirements.
A constraint solver (Answer Set Programming) generates possible mechanics in a given game domain according to hard (required) or soft (optimized) design requirements.
An AI planner tests playability by using generated mechanics to prove that designer-specified requirements for good gameplay can be achieved with the mechanics.
For example, players must be able to reach the end of a level or win a battle without dying.
Unlike AI planners that solve game levels using a fixed set of operators (mechanics), mechanic generation creates the operators.
A planning operator representation supports our goals for a composable and domain-independent mechanic representation.

Together, the constraint solver and planner can generate or adapt game mechanics in a (relatively) domain-agnostic fashion while ensuring the mechanics achieve desired play experiences.
We demonstrate our mechanic generation system in three game domains: platformer game movement mechanics, role-playing game (RPG) spell systems, and a domain combining the platformer and RPG domains.

%%%%%%%%%%%%%%%%%%%%%%%%%%%%%%%%%%%%%%%%%%%%%%%%%

\section{Related Work}

% game generation
Game generation systems take a human-specified set of possible domain content and synthesize possible game mechanics (and game content).
%Two major strands of research have investigated board game and arcade game generation.
%Board game generation systems have coupled a grammar-like description language for the domain to an evolutionary algorithm with fitness functions based on criteria for game playouts (playtraces) and game structure \cite{pell1992:metagame,hom2007:gamegen,browne2010:ludi}.
%In some cases these have led to commercially published products, but have remained tightly tethered to domain-specific representations and evaluation criteria.
%
Game generation in arcade games (similar to the games we model) has focused on assigning collision and movement logic to game entities using predefined tables enumerating possible choices.
Researchers have used evolutionary search \cite{togelius2008:gamegen}, constraint satisfaction \cite{smith2010:variations}, and rule-based systems \cite{treanor2012:gameomatic} to generate games meeting soft optimization criteria and/or hard constraints.
Rather than use top-down enumeration of game mechanics, \citeauthor{cook2013:mechanicminer} Mechanic Miner \cite{cook2013:mechanicminer} used a bottom-up approach---program reflection---to manipulate game mechanics by changing the values used in program functions.
We use a top-down mechanic representation that supports \textit{de novo} synthesis of mechanics and several (relatively) domain-independent evaluation criteria.

Complementing game generation systems, Nelson and Mateas argued for recombinable mechanics to support human designers \cite{nelson2008:recombn}.
Researchers have supported human-defined mechanic analysis with playability checking, using 
simulations in Petri net models \cite{dormans2009:machinations}, 
model-checking and proof in extensions of the event calculus \cite{smith2010:ludocore,smith2013:quantify-play}, 
and simulations or model-checking in other action languages \cite{osborn2013:gamelan}.
Our system draws on related logical models (planning) and further can can generate games using the mechanics being modeled.

% the difference
%Our mechanic model is a symbolic game model that allows verification of game properties through proofs or executions in the symbolic language.
%We focus on avatar-based mechanics, similar to Sicart's \shortcite{sicart2008:defining-mechanics} notion of game mechanics as functions called by game agents.
%Unlike game generation efforts to date, we allow low-level, cross-domain composition of game mechanics.
%As a result, our mechanic language is more limited in expressivity to support automated search for game mechanics.

% GDLs
Mechanic generation and game description languages (GDLs) share a concern for composable mechanic representations.
%Research in GDLs has led to a variety of (primarily) grammar-based generation methods.
The Stanford Game Description Language \cite{love2008:ggp-spec} models turn-based, competitive games in a declarative language and has extensions for randomization and incomplete information \cite{thielscher2010:ggp-spec-incomplete-info}.
A variety of research efforts have modeled specific classes of games using similar (context-free or graph) grammar constructs, including arcade video games \cite{schaul2013:pyvgdl}, card games \cite{font2013:card-game-language}, strategy games \cite{mahlmann2011:strat-unit-gen}, action-adventure games \cite{dormans2012:physics-gen}, and puzzle games \cite{lavelle2013:puzzlescript}.
Grammars are effective for embedding design knowledge into a generating system, but are not readily combined across genres.
We avoid this limitation through a cross-domain mechanic representation that is also amenable to automated generation.

%\subsection{Planning}
Our model of game mechanic structure draws from work on domain representations used in AI planning.
Planning is a process of finding a sequence of operations that transform the world from an initial state into a state in which the goal situation holds.
% PDDL overview
Modern plan representations were developed to scale traditional AI techniques to complex domains by providing additional problem structure knowledge to AI search processes.
Planning representations can often be converted (e.g. to SAT problems) to improve the performance of other approaches through additional representational factoring \cite{russell2009:ai}.
%
%	STRIPS
%		pre, add, del
%		no numerical values
%		goal requirements
%
STRIPS \cite{fikes1972:strips} was one of the earliest planning representation languages, modeling actions in terms of logical predicates.
In STRIPS, operators are a set of preconditions that must hold before the action can be executed and a set of effects that add or delete predicates from the state of the world.

The Planning Domain Description Language (PDDL) \cite{mcdermott1998:pddl} is an ongoing project to extend planning representations to address more complex tasks while building a shared language for research competitions.
%
%	PDDL 1.2
%		allows "neq"
PDDL extended STRIPS-like representations with non-equality constraints, 
%
%	PDDL 2.1
%		adds numeric fluents
%		also: durative/continuous actions (we only use instantaneous); timed initial literals (happen regardless of plan)
%
numeric fluents to model continuous domains, operators with duration, and timed initial literals that modify the world state at fixed times regardless of agent actions.
By modeling mechanics in a similar manner to planning domains we can leverage existing work on planning technologies to check game playability.

%%%%%%%%%%%%%%%%%%%%%%%%%%%%%%%%%%%%%%%%%%%%%%%%%

\section{Mechanic Design Formalization}

In this section we define the mechanic generation problem, provide a model for cross-domain composable game mechanics, and present methods to automatically generate and test game mechanics for given game content.
\textit{Mechanic generation} is the problem of constructing a (set of) game mechanic(s) such that they meet playability requirements to create a desired range of player experiences (allowing and forbidding action sequences) while meeting design requirements on mechanic structure.
\textit{Playability requirements} ensure a game is playable to a given goal, potentially subject to limitations on the states entered or actions taken to achieve the goal.% (e.g. reaching the end of a platformer level without dying and never double-jumping).
\hspace{1pt} \textit{Design requirements} ensure mechanics adhere to designer requirements for how actions work in a game. %(e.g. not having spurious effects or being overly constrained to a single use case).
Both playability and design requirements may be domain-independent or domain-dependent.
Together, design requirements shape mechanics to the form a designer desires while playability requirements ensure those mechanics have desired functions in game content.

Mechanic generation uses a game domain definition to know what may be changed by mechanics.
A \textit{game domain} defines the entities that make up a game, their parameters, and how game states change.
A game domain consists of a \textit{state model}---specifying domain entities, their parameters, and allowed ranges of values---and a \textit{transition model}---specifying how states change from one to another.
In our formulation, the transition model is the set of game mechanics.
The focus of this paper is \textit{avatar-centric} mechanics---transitions initiated by the player (or other in-game agents) in the process of controlling an avatar.
A solution to the mechanic generation problem is a transition model that meets design and playability requirements, given a state model and a set of relevant game instances.
A \textit{game instance} defines a setting from a game domain; e.g., a level in a platformer or a single battle in a role-playing game.
Initial game state is part of a game instance.
Combining a game instance for a state model with a transition model (including avatar-centric mechanics) yields a playable game experience.

As a running example to illustrate our definitions, consider a simple role-playing game (RPG) battle game domain.
RPG battles involve two opposing parties taking turns to attack one another using various spells until one party is slain; the \textit{Dungeons and Dragons} tabletop RPGs are a paradigmatic example.
The RPG state model has a player character and an enemy, each with health and mana resources.
One game instance has the player starting with 3 health and 5 mana (a spell-casting resource) while the enemy has 2 health and 2 mana.
Many variant instances may be considered simultaneously when mechanics are generated.
A playability requirement can ensure that, over all given instances, the player can kill the enemy (reduce enemy health to 0 or less) without being killed.
A design requirement can specify that all spells have a cost---e.g. requiring that every spell cost the avatar that uses it some resource (health or mana).
Mechanic generation asks: given these requirements what spells should be in the game?

A system that solves the mechanic generation problem requires a state and transition model representation, a process to search for transition models that meet design criteria within the state model, and a process to test that transition models meet playability criteria across a provided set of game instances (potentially all valid game instances in that space).
Below we present our state and transition models to generate avatar-centric mechanics as planning operators.
%Below we present our state and transition models appropriate to avatar-centric mechanic generation.
%Our transition model encodes mechanics as planning operators.
Planning operators are a natural representation for game mechanics as operators were designed to represent domains involving sequential choices while readily allowing composition of operator preconditions and effects.
Our operator representation can directly reference variables in the game engine to operate at a very low level of primitives.
We use a constraint solver to search for a transition models (mechanics) that meet design criteria.
Constraint solvers are a valuable generic approach to search combinatorial spaces that readily encode hard and soft requirements on solutions \cite{smith2011:asp-pcg}.
We use a planner to validate transition models against playability criteria.
Planners are effective for proving the presence or absence of play traces (sequences of mechanic choices and state updates) within a game domain.

\subsection{State Model}
\label{sec:state-model}

Our representation for game domains uses a subset of the ideas used in PDDL with extensions specific to games.
Currently we focus on turn-based domains with deterministic actions to simplify our early exploratory work.
The state model defines a domain of game entities in terms of their allowed states as these are core to modeling avatar-centric mechanics (state transitions).

The state model is a set of terms defining entities, parameters, and allowed parameter value ranges for entity parameters in the game world ($AbsRange$) or mechanic changes to those values ($RelRange$).
Terms have the following forms:
\begin{quoting}[vskip=2pt]
\centering
$Entity(e)$ \hfill $Parameter(p)$ \hfill $Has(e,p)$ \\
$AbsRange(p, e, r)$ \hfill $RelRange(p, e, r)$
\end{quoting}
where $e$ is a symbol representing an entity, $p$ is a parameter of an entity, and $r$ is a range of values, which may be discrete or continuous.
For simplicity parameters currently range over integer values.
%A game engine using this formalism must expose an API with `get' and `set' methods for game engine variables corresponding to these logical terms.

Referring back to our example RPG spell system we might define the player:
\begin{quoting}[vskip=2pt]
\centering
$Entity(Player)$ \\ $Parameter(Health)$ \hfill $Parameter(Mana)$ \\
$Has(Player,Health)$ \hfill $Has(Player,Mana)$ \\
$AbsRange(Health, Player, [0,3])$ 
$AbsRange(Mana, Player, [0,5])$
\end{quoting}
\noindent $RelRange$ relates to the transition model described in the next section.

%\begin{table}[tb]
%\small
%\centering
%\begin{tabular}{cc}
%\hline
%\textit{Entity(player)}                             & \textit{}                 \\
%\textit{Parameter(health)}                          & \textit{Parameter(Mana)}  \\
%\textit{Has(player,health)}                         & \textit{Has(player,Mana)} \\
%\textit{AbsRange(player,health,{[}0,3{]})}          & \textit{}                 \\
%\multicolumn{1}{l}{AbsRange(player,Mana,{[}0,5{]})} & \multicolumn{1}{l}{}      \\ \hline
%\end{tabular}
%\label{tab:rpg}
%\end{table}

Game instances give concrete settings for state model entities and parameters.
%define game state in terms of the state model terms; typically as the initial state of a game.
We use fluents to represent these values in our planning model, allowing states to change according to the transition model.
In our RPG example, we can set player health to initially be $3$ using $Initial(Health(Player), 3)$
where $Initial$ sets entity parameter values that hold at the beginning of the game.
%Note the use of single values rather than ranges when initializing instances from a state model.

\subsection{Mechanic Model}
\label{sec:mech-model}

A set of mechanics define a transition model that allows forward simulation and playability checks as planning.
%The transition model is a set of mechanics that allows forward simulation (making results playable as simple text-based games) and playability checks as planning.
%
Consider modeling an RPG spell that causes damage over a period of time.
Such a spell needs to specify several things: conditions on when the spell may apply (e.g. not affecting dead characters), how much damage is done, and at what time(s) the damage is done.
To address examples like this we have drawn from PDDL's action schemas to define an avatar-centric mechanic as a tuple: $\langle i, P, E \rangle$
where $i$ is a unique identifier for a mechanic, $P$ is a set of the preconditions needed for mechanics to occur, and $E$ is a set of effects of performing the mechanic.

Our preconditions and effects extend traditional PDDL action schemas with time-indexing and coordinate frames of reference.
Time-indexing allows preconditions to reference state at times other than the present and allows effects to reference states other than the next game state.
Games often incorporate delayed effects or checks on historical state, motivating our time-indexing extension.
Coordinate frames distinguish between traditional world-state terms and ``perceived'' avatar-relative versions of world terms.
Absolute frames of reference model requirements on the state of the world.
Relative frames of reference capture the intuitive notion that many avatar-centric game mechanics have preconditions and effects relative to an avatar, rather than absolute world state (e.g. adjacency as relative position).

\begin{sloppypar}
Our planning model implements semantics for a subset of PDDL with extensions appropriate to our definition.
$AbsRange$ is used to specify valid absolute frame of reference values while $RelRange$ is used for relative frames of reference.
Preconditions test game state; we allow tests for equality, inequality, and lesser-than and greater-than relations.
All preconditions and effects are tuples of the form $\langle frame, time, condition \rangle$; where $frame$ indicates a coordinate frame of reference, $time$ specifies a time-index, and $condition$ specifies a game state value to check for (or update).
In our formalism, a condition takes the form $F(parameter(entity), value)$ where $F$ is a logical function that either tests two values and returns a boolean value (for preconditions) or updates an entity parameter value (for effects).
Testing for the avatar currently being alive would be $\langle Absolute, 0, GreaterThan(Health(Player), 0) \rangle$.
\end{sloppypar}

Effects update game state.
%Effects may reference current or future state; we forbid historical changes (time-travel paradoxes may result).
For absolute frames of reference updates set state to a particular value (constrained within $AbsRange$); for relative frames of reference updates change state values by a given amount (constrained within $RelRange$).
%Effects take the same form as preconditions and are interpreted as logical rules for updates or setting appropriately.
A spell that checks for the enemy being alive and reduces enemy health by 1 on the two next turns is:
\begin{minipage}{\linewidth}
%\small
\begin{tabbing}
$\langle Da$\=$mageOverTime$, \\
\>$\{ \langle$\=$Absolute, 0, GreaterThan(Health(Enemy), 0) \rangle \}$, \\
\>$\{ $\= $\langle Relative, 1, Update(Health(Enemy), -1)\rangle $, \\
\>\> $ \langle Relative, 2, Update(Health(Enemy), -1)\rangle \} \rangle $
\end{tabbing}
\end{minipage}

Mechanic \textit{recombination} occurs when one mechanic references another mechanic having occurred.
Fighting game or rhythm game combo systems exemplify avatar-centric recombination.
Mechanic recombination naturally encodes event-relevant mechanics, rather than being limited to mechanics that reference state.
For mechanic recombination we allow preconditions and effects to reference the \textit{event} of a mechanic occurring with $Performed(i)$.
Semantically, a mechanic as a precondition requires that mechanic to have (or not have) occurred at a time index.
For example, a double-jump may require a player to have jumped at the previous time-step: $\langle Absolute, -1, Equal(Performed(Jump), Player)\rangle$ 
When $Performed(i)$ appears as an effect the preconditions and effects of that mechanic are applied.
The mechanic using $Performed(i)$ as an effect indicates the time to apply the performed mechanic.
Note that frames of reference are not relevant for mechanic indexes (these are provided by the indexed mechanics themselves) and are ignored.

As in PDDL, we assume inertial state and circumscription: any entity parameter not affected by a mechanic continues to hold its previous value.
$Performed(i)$, however, is treated as an \textit{event} and not subject to inertial state.

%%%%%%%%%%%%%%%%%%%%%%%%%%%%%%%%%%%%%%%%%%%%%%%%%%%%%%%%%%%%%%

\section{Mechanic Generation}

Mechanic generation creates a set of mechanics within a game domain subject to playability and design requirements.
We use a constraint solver to search for a set of mechanics constrained to meet the given design requirements.% while satisfying playability requirements using a constraint solver.
\hspace{1pt}
Design requirements help avoid low-quality mechanic solutions.
Hard design requirements (as used by \citeauthor{smith2011:asp-pcg} \shortcite{smith2011:asp-pcg}) enforce conditions on the form of mechanics or relations among a set of mechanics---e.g. not allowing a mechanic to have both equality and non-equality preconditions for the same game state or requiring no two mechanics to have identical preconditions and effects.
Soft design requirements (as reviewed by \cite{togelius2011:sbpcg}) give optimization criteria for what makes (sets of) mechanics better or worse---e.g. aiming to minimize the number of preconditions and effects used by a mechanic in favor of simplicity.
Playability is evaluated using a planner (described in the next section) to prove a player can meet playability requirements on given test game instances.

We use Answer Set Programming (ASP) \cite{baral2003:asp}---a form of declarative programming---to implement the constraint solver and planner (see also \cite{gebser2010:coala-asp}).
In our ASP implementation, preconditions and effects index the mechanic they are part of.
Hard requirements are integrity constraints on combinations of preconditions and effects; soft requirements optimize over weighted and prioritized sums of terms.
Hard and soft requirements may refer to to parts of one or more mechanics.
Implementing both the generation and testing components of our system in ASP allows a single monolithic synthesis process; either part may be run independently (or use other systems such as GGP players \cite{love2008:ggp-spec}).

Mechanic generation creates mechanics by choosing preconditions and effects for each mechanic while ensuring the mechanics conform to design requirements.
Some design requirements apply across types of games (not requiring a state hold and not hold at the same time) while others are more domain-specific (spells should be ``balanced'' in terms of resource costs to execute vs effects on avatars).
Given a set of operators, a planner proves whether a plan exists for given game content subject to playability requirements.
The process of generating mechanics using a constraint solver and testing those mechanics with a planner repeats until all hard requirements are met and all soft requirements are optimized.
While this is an expensive process we have started with small game domains to explore the relevant research problems.
Note also that many games use relatively small sets of mechanics (e.g. RPG spell systems, platformer movement mechanics, card game rules, etc.).

\subsection{Playability Checking}
\label{sec:playability-check}

We use a simple planner that proves that playability requirements can be met in game instances with a given set of mechanics.
The planner uses playability requirements as goal situations to prove whether a plan exists that can meet playability requirements.
For convenience we use ASP as our implementation language for the planner.

Playability requirements come in three forms: (1)~goals, (2)~maintenance goals, and (3)~engine constraints.
Goals give a game agent target situations to seek; the planner must prove the presence of a plan that meets the goal.
Maintenance goals give situations that begin true and must hold throughout the plan (e.g. being alive); the planner must prove a plan achieving the goals always upholds maintenance goals.
Maintenance goals are useful for specifying failure criteria in a game as the negation of a failure state must always hold.
Engine constraints enforce semantics mapping to non-avatar rules in a game engine (e.g. preventing two entities from occupying the same space); the planner must follow these constraints when making plans.
In our RPG battle example, the player goal is to kill all enemies while maintaining the state of being alive (not being killed) and an engine constraint ensures the player cannot drop below 0 mana.

Building our model off planning domain representations provides a simple, factored logical model of the game world that affords game mechanic combination and synthesis while also yielding playable games.
The planner can forward simulate a game instance using player choices among generated mechanics and enforce player victory or failure based on goals and maintenance goals.
Game state, allowed operators, and goals can be presented through a simple text-based interface.

%	future
%		conditional effects ("when" effects)
%		range equality
%		inequality
%		state constraints over time periods (state-trajectory constraints)
%		basic type system
%		probabilistic effects (PPDDL)

\subsection{Mechanic Adaptation}
Instead of generating mechanics from scratch, \textit{mechanic adaptation} starts with a set of mechanics and produces a minimally changed set of mechanics.
Mechanic adaptation uses mechanic generation for iterative design.
In iterative design a set of mechanics are tested and adjusted to meet new insights about the game---adaptation requirements.
Mechanic adaptation is given the same inputs as mechanic generation along with an initial set of mechanics and new adaptation requirements.
\textit{Adaptation requirements} specify additional playability or design requirements for mechanic generation.
New playability requirements may indicate additional goal states for the player to pursue or identify unwanted states.
New design requirements may control the amount of change to make to a set of mechanics.
The definition of `minimal change' varies by game domain and must be specified to adapt mechanics.

Mechanic adaptation takes the same input game state and transition models as mechanic generation augmented with a pre-existing set of game mechanics.
Adaptation adds or removes preconditions and effects from existing mechanics and may also generate new mechanics.
Changes to mechanics must meet designer-specified criteria for minimality while adhering to all adaptation requirements.
We adapt mechanics by having the constraint solver perform the standard generation process but seeded with the additional mechanics.
The previous set of design requirements are given along with new adaptation requirements and a definition of minimality (e.g. minimizing the total number of changes made).
Mechanic adaptation performs the same loop of generating and testing possible mechanics as in mechanic generation.

%%%%%%%%%%%%%%%%%%%%%%%%%%%%%%%%%%%%%%%%%%%%%%%%%

\section{Examples}

Our game domain formalism supports a variety of avatar-centric mechanic systems.
In this section we illustrate how to represent a simple role-playing game (RPG), a simple platformer, and a game that merges these two systems.
RPGs require a balanced and diverse set of character spells.
Platformers are games where a character navigates physical obstacles in a virtual space, exemplified by the \textit{Super Mario Bros.} games.
Platformers require a finely tuned and widely reused small set of spatial navigation mechanics.
We generate spells in the RPG and movement mechanics in the platformer.
By concatenating these two domains we illustrate how our model affords cross-domain mechanic generation.

%%% RPG %%%
\subsection{Role-Playing Game}
We define RPG combat mechanics using a set of entity attributes and resources (here health and mana for the player and a set of enemies) as above.
%Our earlier RPG spell example defines this basic domain.
Playability requirements give: a player goal situation of having all enemies dead, a player maintenance goal of not being dead; and an engine constraint preventing negative mana.
Together, these playability requirements encode the basic notion of an RPG battle as killing an opponent without being killed while having bounded resources.
Two domain-independent design requirements give: a hard requirement to prevent mechanics from having preconditions that force a predicate to equal more than one value and a soft requirement to minimize the number of preconditions and effects of mechanics to produce the simplest set of mechanics.
Many domains have a notion of actions having costs; our domain-specific version of costs requires all actions incur a mana or health cost.

Our system generated a variety of RPG spells using the game domain, a game instance with two enemies, and the playability and design requirements above.
Playtraces are plans: a series of entity actions (spells used) that inflict damage and cost health or mana.
One example spell was given above, others typically have simple effects such as inflicting damage at a single time point or affecting multiple targets:
\begin{minipage}{\linewidth}
%\small
\begin{tabbing}
$\langle $\=$DamageAll$, $\{\}$, \\
\>$\{ $\= $\langle Relative, 1, Update(Health(Enemy1), -1)\rangle $, \\
\>\> $ \langle Relative, 1, Update(Health(Enemy2), -1)\rangle $, \\
\>\> $ \langle Relative, 1, Update(Mana(Player), -2)\rangle \} \rangle $
\end{tabbing}
\end{minipage}

\noindent where there are no preconditions and the effects damage both enemies while costing the player mana.
Note that we have given human-readable names to the mechanics; internally $i$ (the name) is an integer.
Also note that our examples were chosen to illustrate the most semantically sensible mechanics generated; by definition all mechanics achieve playability and design requirements.

%%% platformer %%%
\subsection{Platformer}
%Platformers can be described in terms of a set of entities (here the player, blocks, and enemies) each assigned spatial coordinates (here two values corresponding to two spatial dimensions).

%\begin{quoting}[vskip=1pt]
%\centering
%$Entity(player)$ \\
%$Parameter(xPos)$ \hfill $Parameter(Ypos)$ \\
%$Has(player,xPos)$ \hfill $Has(player,Ypos)$ \\
%$AbsRange(player, xPos, [1,8])$ \\ $AbsRange(player, Ypos, [1,6])$
%\end{quoting}
%The initial state of the player for our example (Figure~\ref{fig:platformer}) is $Initial(xPos(player), 1)$, $Initial(Ypos(player), 2)$.

\begin{figure}[tb]
\centering
\includegraphics[width=0.8\linewidth]{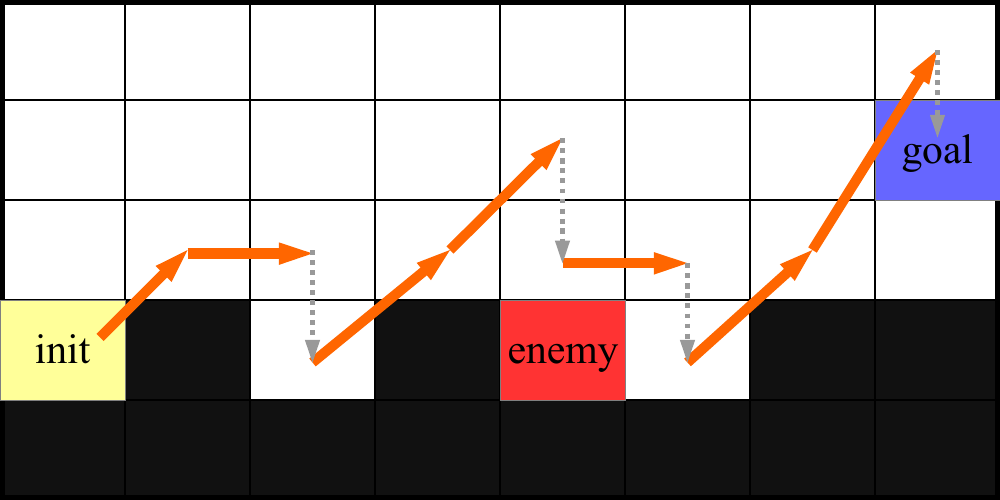}
\caption{
Platformer level showing a playtrace using a generated mechanic set.
Arrows indicate generated mechanics, dotted arrows indicate gravity.
}
\label{fig:platformer}
\end{figure}

We define two-dimensional platformers as a set of entities (here the player, blocks, and enemies) each assigned spatial coordinates (Figure~\ref{fig:platformer}).
The platformer has playability requirements for: a player goal situation of reaching the end, a player maintenance goal of not overlapping with an enemy; and an engine constraint preventing the overlap of any entity and a block.
Another engine constraint enforces gravity by requiring all entities to move down one unit each turn if that space is not occupied by a block.
We reused two design requirements from the RPG example: preventing exclusive pre-conditions and minimizing the number of mechanic preconditions and effects.
A third soft requirement optimizes for as few mechanics as possible (to create a ``tighter'' game system) and a fourth soft requirement minimizes the number of different entities referenced by mechanics (favoring motion of a single avatar).

Figure~\ref{fig:platformer} illustrates a simple platformer level and shows one trace found by the planner that moves the player avatar to the goal position.
The planner generated mechanics for moving forward, jumping, and double-jumping (indicated by arrows).
Dotted arrows indicate the effects of gravity.
$DoubleJump$ illustrates an event precondition requiring $Jump$ to have occurred immediately before:
\begin{minipage}{\linewidth}
%\small
\begin{tabbing}
$\langle$\=$Jump$, \\
\>$\{ $\= $\langle Relative, 1, Equal(Ypos(e), Ypos(Block)$+$1) \rangle $, \\
\>\> $\langle Relative, 1,  Equal(Xpos(e), Xpos(Block)) \rangle \} $, \\
\>$\{ $ \> $\langle Relative, 1, Update(Xpos(e), 1) \rangle $, \\
\>\> $\langle Relative, 1, Update(Ypos(e), 1) \rangle \} \rangle$ \\
%\\
$\langle$\=$DoubleJump$, \\
\>$\{ $\= $\langle Relative, 1, Equal(Ypos(Player), Ypos(Block)$+$1) \rangle $, \\
\>\> $\langle Relative, 1,  Equal(Xpos(Player), Xpos(Block)) \rangle $, \\
\>\> $\langle Absolute, -1, Equal(Performed(Jump), Player) \rangle \}$, \\
\>$\{ $ \> $\langle Relative, 1, Update(Xpos(Player), 1) \rangle $, \\
\>\>$\langle Relative, 1, Update(Ypos(Player), 2) \rangle \} \rangle$ \\
\end{tabbing}
\end{minipage}

%%% platformer-RPG %%%
\subsection{Combined Game}
To test the modularity of our representation we concatenated the previous two domains to create a ``platformer-RPG'' game.
All game state definitions are unchanged: combining RPG resources and platformer location only makes entity state more complex.
We retain the previous playability requirements from both domains with conjunctive (all criteria must be met) goals, maintenance goals, and engine requirements.
With these simple changes we can generate mechanics appropriate to the domain such as attacking at a distance with a spell:
\begin{minipage}{\linewidth}
\begin{tabbing}
$\langle Magic$\=$Missile$, \\
\> $\{$\=$ \langle Relative, 0, Equal(Xpos(Enemy), 2) \rangle$, \\
\>\> $\langle Relative, 0, Equal(Ypos(Enemy), 0) \rangle \}$, \\
\> $\{$\>$\langle Relative, 0, Update(Health(Enemy), -1) \rangle \} \rangle$
\end{tabbing}
\end{minipage}
where the preconditions check for an enemy two spaces in front of the player and the effect reduces enemy health.

\section{Richer AI Design}
To further develop the design tasks in the previous example domains we extended our system to generate mechanics for multilevel progressions, multiagent games, and map controls to mechanics.
These additions illustrate how our representation can model some more complex design tasks that directly relate to mechanics.

\subsection{Multilevel Progression}
Platformers (and most game genres) often gradually introduce new mechanics to players over a sequence of levels.
Generalizing mechanic generation to include requirements on which mechanics are used along a progression requires two additions: planning across multiple levels and providing requirements on mechanic use.
To implement multilevel progression we augmented the initial state and playability requirement definitions with a level index of the form $Initial(level, paramter(entity), value)$.
Playability checks must ensure the given mechanic set can yield valid playtraces for all levels provided, treating each as a separate planning problem with the same set of mechanics.

The constraint solver can enforce various notions of progression across multiple levels.
For example, we have required an increasing number of mechanics be used in each level over a level progression.
We have also required that the specific mechanics used in each level reappear in all subsequent levels.
By using increasingly complex game instances the generated mechanic sequences can introduce weaker and stronger (larger effect) versions of the same mechanic.
This has produced the $DoubleJump$ mechanic above.
These progression requirements encode a notion of training players by needing to master additional skills (c.f. \citeauthor{butler2013:progression-tool} \shortcite{butler2013:progression-tool}; \citeauthor{dormans2010:missions} \shortcite{dormans2010:missions}; \citeauthor{andersen2013:trace} \shortcite{andersen2013:trace}).
We have used our atomic representation to require the progressive introduction of preconditions or effects (as in the $DoubleJump$ introduction of an event precondition).

\subsection{Multiagent Games}
RPG battles typically involve competing agents.
To incorporate multiagent modeling we augmented our planner to track actions and perceived state relative to each agent.
We now indicate agent-specific goals and maintenance goals (engine constraints are currently treated as universal).

Playability checks optimize toward all agent (potentially competing) goals.
To ensure plans are possible we typically require that the player can achieve her goal situation before any opposition, but that both goal situations can be achieved within a prescribed number of plan steps.
Alternatively, we have also provided goals that are intended to improve player experience without directly negating the player's maintenance goals (e.g. trying to minimize player health, rather than kill the player).
Adding multiagent modeling is computationally costly but allows broader modeling of competition (or collaboration) interactions.
True adversarial agent interactions, however, will require a more sophisticated planner or adoption of general-purpose adversarial game players.

\subsection{Controls}
Platformers depend heavily on the game controls.
Our modular representation can readily map a given set of input buttons to generated mechanics.
We define the input commands, add these controls as additional preconditions for mechanics, and require there is always a single unambiguous mechanic for an input.
Hard design requirements state that all mechanics have at least one input and no two mechanics with the same preconditions use the same set of inputs.
Additional soft design requirements encode a simple notion of ``intuitive'' mappings by maximizing the use of overlapping sets of buttons for mechanics with effects on overlapping sets of entity-parameter states.
Automated control generation can make a game playable on different game platforms (e.g. via mobile phone touch screen or game controller) though this will require more detailed representations of control ``feel'' \cite{swink2009:gamefeel}.

%While we have not yet addressed more complex control mapping problems than these (such as mapping to real-world physics or durative button presses \cite{swink2009:gamefeel}) the ability to add these capabilities readily to our  shows the promise of our formalism.
%Directly encoding controls for mechanics further opens the possibility to model agents that must use a control set (with the associated design complications), rather than directly triggering mechanics.
%Using agents working from controls, we plan to model how control sets influence player interaction.

% future questions:
%	integrate domains
%	more sophisticated aspects of gameplay - not just minimal path, but more general trace-based or state-based features => what makes gameplay good beyond being able to win/lose?
%	mechanics including new game state features

%%%%%%%%%%%%%%%%%%%%%%%%%%%%%%%%%%%%%%%%%%%%%%%%%

\section{Conclusions}

In this paper we formalized the mechanic design problem, presented a domain-independent representation for avatar-centric mechanics, and illustrated how to generate and combine mechanics using a constraint solver and planner.
By using a domain-independent representation our system can readily work in a variety of game domains, focusing on the higher-level problems of designing mechanics rather than genre-specific concerns.
%We plan to extend these design reasoning capabilities to more complex design tasks such as adding new parameters or entities to a game domain to accompany new mechanics---e.g. adding food and a hunger system to a RPG.
%Having the system provide design justifications beyond meeting playability requirements will be important to guide the system toward interesting, human-playable games.
Developing more sophisticated playability requirements---such as reasoning on trajectories of actions or states---can support a broader class of design concerns.
Further, reasoning over the space of potential gameplay outcomes will be needed to control for \textit{expected} gameplay outcomes.
Formally modeling these tasks provides insight into challenges in the game design process.
Autonomous mechanic generation (given designer initial inputs) holds promise for creating AI designers that generate games starting from mechanics.
%Game generators can provide games for a variety of applications spanning educational assessment and teaching through citizen science and data collection.

% learning
% transfer
% sketching games
% player models
% domain integration
% skinning

%\pagebreak

\bibliography{lib}
\bibliographystyle{aaai}

\end{document}